\documentclass[10pt,letterpaper]{article}

\usepackage{cogsci}

\cogscifinalcopy % Uncomment this line for the final submission 

\usepackage[export]{adjustbox}
\usepackage{pslatex}
\usepackage{apacite}
\usepackage{float} % Roger Levy added this and changed figure/table
\usepackage{amsmath,amsfonts, amssymb, mathrsfs }
\usepackage{graphicx}
\usepackage{enumitem}

\newcommand{\SRSA}{S_\text{RSA}}
\newcommand{\SLit}{S_\text{Lit}} 

\newcommand{\word}[1]{\textit{#1}}

\newcommand{\R}{\mathbb{R}}

\newcommand{\U}{\mathcal{U}}
\renewcommand{\L}{\mathcal{L}}

\usepackage{xcolor}

\title{Color Overmodification Emerges from \\ Data-Driven Learning and Pragmatic Reasoning}

\author{
  {\large \bf Fei Fang$^{1}$} \\\And
  {\large \bf Kunal Sinha$^{2}$} \\\And
  {\large \bf Noah D.~Goodman$^{3,1}$} \\[1ex]
  $^1$Department of Computer Science\quad$^2$Department of Symbolic Systems \\ $^3$Department of Psychology\quad$^4$Department of Linguistics\\
  Stanford University \\
  Stanford, CA 94305 USA \\
  \texttt{\{feifang, ksinha2, ngoodman, cgpotts, ekreiss\}@stanford.edu} \\\And
  {\large \bf Christopher Potts$^{4}$} \\\And
  {\large \bf Elisa Kreiss$^{4}$} \\
}

\begin{document}

\maketitle
\begin{abstract}
Speakers' referential expressions often depart from communicative ideals in ways that help illuminate the nature of pragmatic language use. Patterns of \emph{overmodification}, in which a speaker uses a modifier that is redundant given their communicative goal, have proven especially informative in this regard. It seems likely that these patterns are shaped by the environment a speaker is exposed to in complex ways. Unfortunately, systematically manipulating these factors during human language acquisition is impossible. In this paper, we propose to address this limitation by adopting neural networks (NN) as learning agents. By systematically varying the environments in which these agents are trained, while keeping the NN architecture constant, we show that overmodification is more likely with environmental features that are infrequent or salient. We show that these findings emerge naturally in the context of a probabilistic model of pragmatic communication.

\end{abstract}

\section{Introduction}

\begin{figure}
    \centering
    \includegraphics[width=1\linewidth]{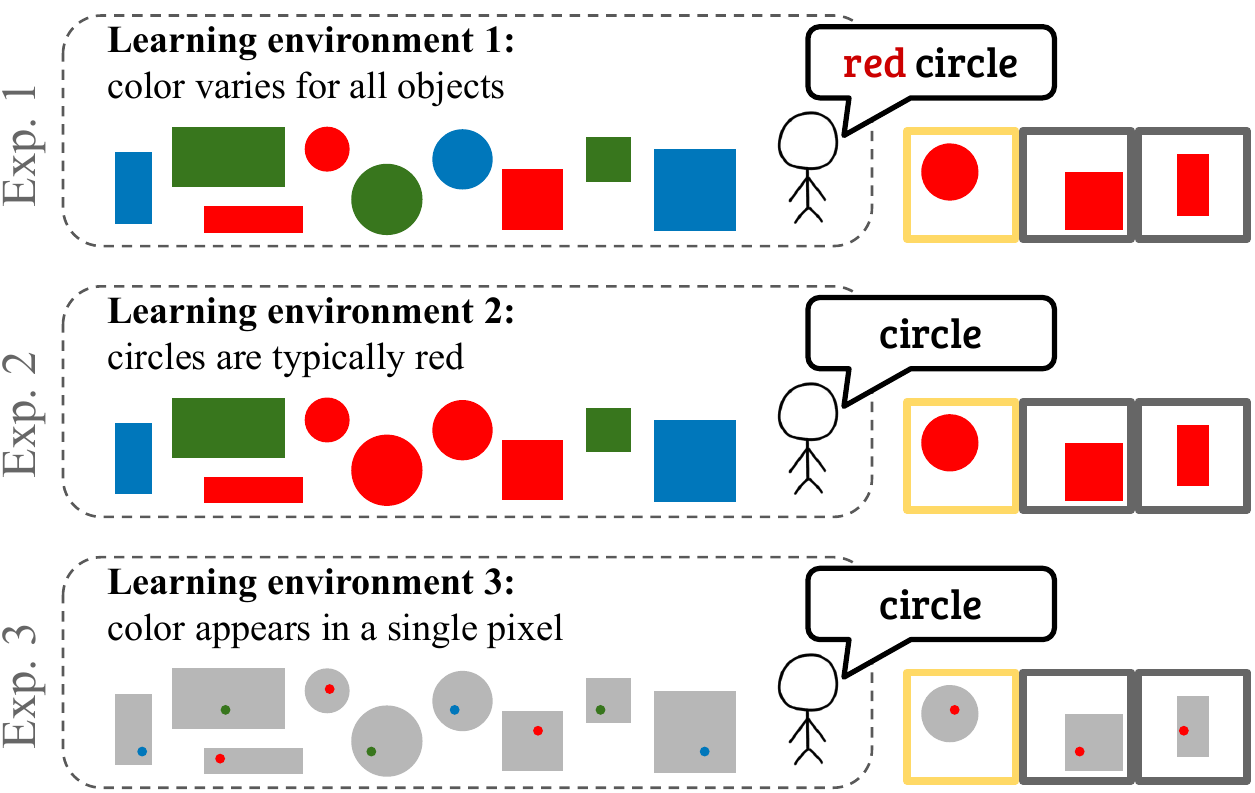}
    \caption{Neural speakers are exposed to different environments during learning and tested on their choice of color overmodification. Varying the underlying data distribution shapes neural speakers' overinformative usage of color, explaining previous findings from human production and suggesting additional factors that may give rise to overmodification.}
    \label{fig:reference_game}
\end{figure}

Overmodification describes the well-attested tendency of speakers to supply more information than apparently needed for a listener to identify the intended referent. Consider the example in Figure~\ref{fig:reference_game} (top). The speaker's goal is to communicate the target image (here highlighted with a yellow box) to a listener. While the utterance ``circle'' would suffice for the listener to pick out the target from the context, speakers often use ``red circle'' instead, whereas the unmodified noun ``circle'' is more likely when circles are typically red (middle) or when color in general is not as salient (bottom). In this paper, we show that such patterns arise naturally in neural network speaker agents that are (1) trained in these different environments and (2) reason pragmatically about a listener.

The rate of overmodification with color terms has been shown to vary in structured ways \shortcite{pechmann1989overspecification,engelhardt2006speakers,koolen2011factors}. Non-color-diagnostic objects, which occur in a variety of colors equally frequently (e.g., cups), are readily overmodified \shortcite{sedivy2003pragmatic,rubio2016redundant}. For objects that are associated with one particular color (e.g., bananas), the overmodification rate drops when the color is typical (``banana'' for a yellow banana), and increases when it is atypical (``blue banana'' for a blue banana) \shortcite{westerbeek2015typicality,degen2020redundancy,sedivy2003pragmatic,kreiss2020production}.

In addition to color diagnosticity, the visual salience of color has also been argued to play an important role in why speakers tend to frequently overmodify with color but not with other modifiers such as size \shortcite{sedivy2003pragmatic,van2019conceptualization,davies2013speakers,rubio2019contrastive,rubio2016redundant,brown2011experimental}. This intuition predicts that overmodification rates decrease with reduced visual salience.

A variety of computational models have been proposed to account for overmodification patterns. The Incremental algorithm \shortcite{dale1995computational} predicts color overmodification whenever it serves to disambiguate from at least one distractor. The probabilistic referential overspecification model (PRO; \shortciteauthor{van2019conceptualization}, \citeyearNP{van2019conceptualization}) predicts that speaker preferences and visual salience explain the overmodification patterns by arguing that atypically colored objects are more salient. Recently, several models of overmodification have been proposed within the Rational Speech Acts (RSA) framework \shortcite{frank2012predicting,goodman2016rsa}, in which overmodification arises from abstract speaker agents reasoning about internal listeners \shortcite{cohn2019incremental,WaldonDegen2021}.
\shortciteA{degen2020redundancy} propose an RSA-based model that predicts overmodification patterns by defining pragmatic reasoning over non-deterministic semantics. In this way, the typicality effect, for instance, arises from the perception that the utterance ``banana'' applies less well to blue bananas than yellow ones.

What could lead a language learner to adopt the sort of semantics that produces such overmodification patterns? 
All previously proposed models assume that the speaker is already equipped with a complete semantics, abstracting away the process through which an agent extracts attributes of an object from raw visual input.
These models therefore cannot provide a perspective on how learning to use language in the world might give rise to such a semantics in the first place. We overcome this limitation by introducing a neural network that learns the semantics from data consisting of raw visual and linguistic input. This allows us to show how overmodification arises directly from the underlying data distribution. 

Our neural network-based approach also allows us to explore the role of pragmatic reasoning in overmodification. To do this, we evaluate a neural pragmatic speaker model (\emph{RSA speaker}) and a neural literal speaker model (\emph{literal speaker}) while varying the underlying data distributions they learn from. Both the literal speaker and the RSA speaker receive the context in the form of images as visual input, and do not observe any systematic overmodification during training. In a series of experiments, we manipulate the data that the models learn from by inducing color typicality (Exp.~2) as well as decreasing the salience of color (Exp.~3), and we compare these results to a baseline distribution (Exp.~1). 
We conclude that both pragmatic reasoning about a listener and properties of the underlying data distribution play a central role in achieving human-like rates of color overmodification. This provides a candidate explanation for the underlying processes giving rise to color overmodification in human language.

\section{Background}
\subsection{Empirical Evidence for Overmodification} \label{sec:language-background}

Overmodification by speakers is a robustly reported phenomenon on referring expression production \shortcite{pechmann1989overspecification,nadig2002evidence,sedivy2003pragmatic,engelhardt2006speakers,koolen2011factors,rubio2016redundant,degen2020redundancy,heller2020production}. Color adjectives stand out since their rate of overmodification is generally higher than for other modifiers such as size and material \shortcite{pechmann1989overspecification,sedivy200517,mitchell2013typicality,gatt2011non}. Two main sources have been proposed to explain this. 
First, in contrast to the relative nature of size adjectives, color adjectives are considered absolute since their attribution to an object is (relatively) context-independent \shortcite{kennedy2007vagueness,syrett2009gradable}. Second, color is inherently visually salient and therefore prioritized in production \shortcite{sedivy2003pragmatic,van2019conceptualization,davies2013speakers,rubio2019contrastive,rubio2016redundant,brown2011experimental,long2021contrast,rubio2021color} due to speaker-internal or listener-oriented processes \shortcite{arnold2008reference}.

Even within the domain of color adjectives, overmodification rates vary based on how closely associated the object is with the presented color, i.e., the object's color-diagnosticity. Based on object recognition experiments, \shortciteA{tanaka1999diagnosticity} motivate a distinction of objects with high and low color-diagnosticity. Low- or non-color-diagnostic objects can plausibly occur in many colors (e.g., cups), whereas highly color-diagnostic objects only have a few highly associated colors (e.g., bananas). While speakers often overmodify with color when referring to non-color-diagnostic objects, they rarely do so with color-diagnostic ones that are presented in their typical color \shortcite{sedivy2003pragmatic}. Moreover, the rate of overmodification has been shown to increase linearly with decreasing typicality of the color for the object \shortcite{westerbeek2015typicality,degen2020redundancy}. For instance, a yellow banana would receive the least, a blue banana the most, and a brown banana an intermediate rate of overmodification.

While this literature rigorously explores color overmodification, the question of how changes in the visual environment of a learning agent causally drive overmodification remains elusive to empirical methods. In Exp.~1 and 2, we explore how the general phenomenon of overmodification arises naturally from learning from distinct data distributions. Since these data distributions can be arbitrarily adjusted, they also offer the opportunity to simulate speakers in worlds distinct from ours and study their overmodification patterns (Exp.~3). 

\subsection{Rational Speech Acts (RSA) framework}
The Rational Speech Acts (RSA) framework is a family of computational models of communication \shortcite{frank2012predicting} where two rational agents, a speaker and a listener, are modeled as recursively reasoning about each other when producing and interpreting utterances. This work follows important precedents from game theory \shortcite{Lewis1969convention,jager2007game,franke2009signal,golland2010game}, and has been shown to capture a wide range of linguistic phenomena (e.g., \shortciteauthor{goodman2013knowledge}, \citeyearNP{goodman2013knowledge}; \shortciteauthor{kao2014nonliteral}, \citeyearNP{kao2014nonliteral}; \shortciteauthor{Hawkins:Goodman:2020}, \citeyearNP{Hawkins:Goodman:2020}; \shortciteauthor{van2021probabilistic}, \citeyearNP{van2021probabilistic}).

The probabilistic nature of the RSA framework lends itself naturally to integration with neural network systems. Recent work has shown that RSA reasoning helps in a variety of areas, including continual language adaptation \shortcite{Hawkins:Goodman:2020}, color reference in context \shortcite{monroe2017colors,Monroe:Hu:Jeong:Potts:2018}, following navigational instructions \shortcite{fried-etal-2018-unified}, and context-aware image-based text generation \shortcite{andreas2016reasoning,Vedantam-etal:2017,Cohn-Gordon:Goodman:Potts:2018,nie-etal-2020-pragmatic}.

While previous work has compared the \emph{overall performance} of neural RSA speakers versus non-RSA speakers \shortcite{white2020learning,andreas2016reasoning,monroe2017colors}, we specifically focus on the extent to which such agents rely on overmodification (specifically with color) to communicate. We further extend previous work by investigating how changes in the underlying data distribution that the models learn from affects pragmatic language use. This leads to an account in which overmodification emerges naturally from the visual and distributional properties of the world that a pragmatic speaker is exposed to during learning.

\section{Task}

We explore the production of referring expressions in simple reference games in which a speaker and listener are presented with the same set of three images, varying in order. The speaker's goal is to communicate the identity of the target image to the listener, whose goal is to pick out the target.

\begin{figure}
    \centering
    \includegraphics[scale=0.5]{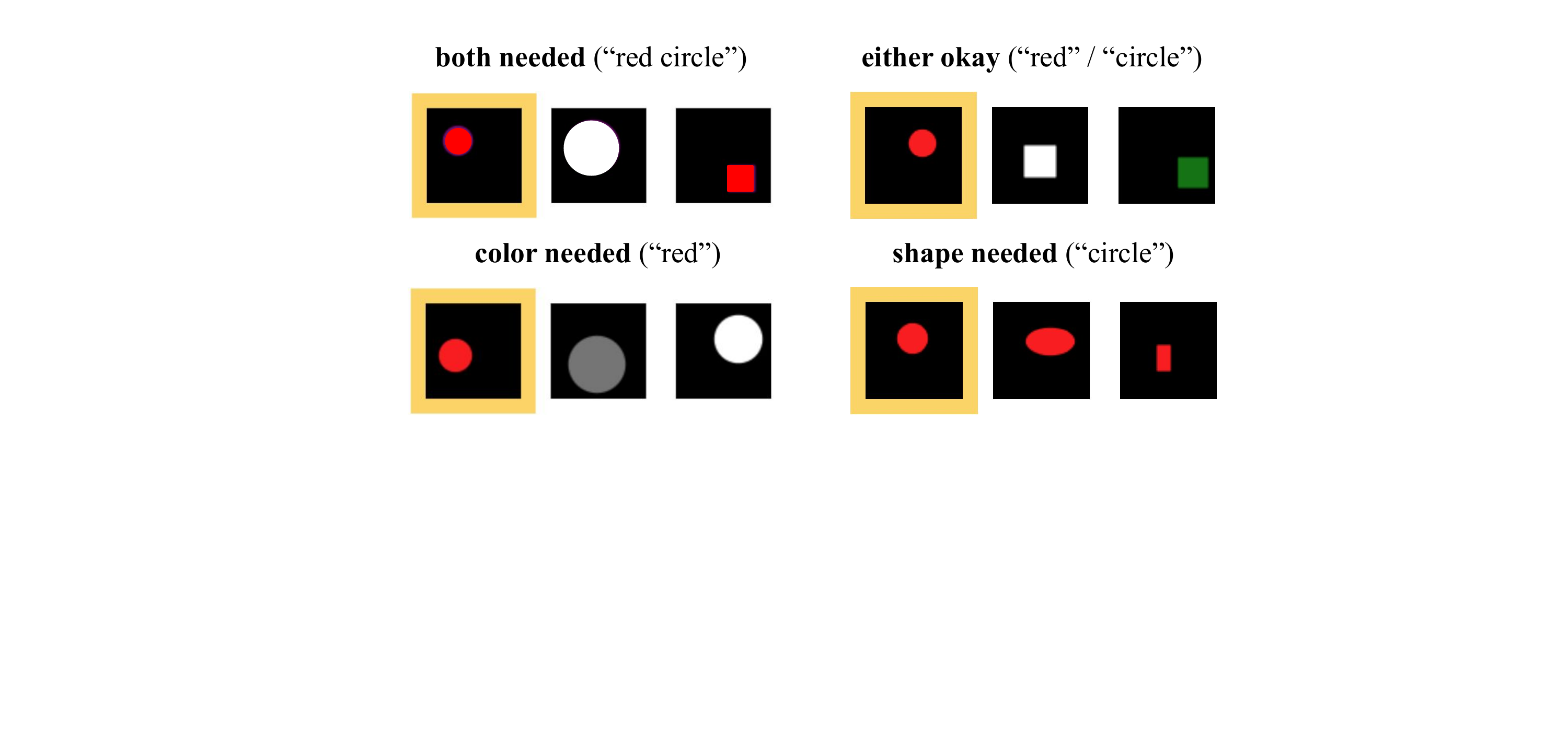}
    \caption{Context conditions, varying in the type of information that a referring expression must provide to distinguish the target. Color overmodification occurs when color is mentioned in the \emph{shape needed} condition.}
    \label{fig:games-by-condition}
\end{figure}

The games are generated synthetically using the ShapeWorld dataset creation framework \shortcite{kuhnle2017shapeworld}. In this setup, six colors are randomly paired with four potential shapes, resulting in 24 unique color--shape compositions. Each generated image displays one colored object that is sized and positioned randomly against a black background. 
Each generated game consists of three distinct images and varies in the minimal amount of information the listener would need to single out the target, resulting in the four \emph{context conditions} exemplified in Figure~\ref{fig:games-by-condition}. Each game is paired with the most concise utterance that establishes unique reference, using a vocabulary of 11 words: those corresponding to the six colors and four shapes, plus the word \word{shape}.

The dataset's synthetic nature allows for highly controlled interventions on inherent properties of the world (e.g., color salience, Exp.~3), which can be difficult if not impossible to simulate with human learners in the real world. 

\section{Models}

We experiment with two different types of speakers, an \emph{RSA speaker} and a \emph{literal speaker with context}. The \textbf{RSA speaker} reasons explicitly about the listener to generate utterances that optimize for successful communication. The \textbf{literal speaker with context} learns to directly map contexts to referring expressions. Based on the utterance produced by a speaker, the \textbf{evaluation listener} chooses one of the referents based on literal semantics learned from non-contextual data.

\paragraph{Evaluation listener}
The listener's goal is to pick out the speaker's intended target referent among three potential referents based on the speaker's utterance.
To do so, the listener $L_\text{Eval}$ constructs a probability distribution over the referents that encodes $L_\text{Eval}$'s belief about which is the intended target. More specifically, given utterance $u$ and referents $r_1, r_2, r_3$, $L_\text{Eval}$ will select referent $r_k$ with probability 
\begin{equation} \label{eq:listener}
   L_\text{Eval}(r_k \mid u, r_1, r_2, r_3) \propto \exp(\L(u, r_k)), 
\end{equation}
where $\L$ is a neural-based semantic function mapping utterances to referents that is learned from data.

Given an utterance $u$ and a referent $r$, $\L$ outputs a probabilistic semantic value $p \in [0,1]$ that represents the listener's judgment about how well $u$ applies to $r$. The probabilistic semantic value is computed as follows: given an utterance $u$ and a referent $r$, \begin{equation}
    \L(u, r) = \sigma(f_L(r)^T g(u)),
\end{equation}
where $f_L$ is an image encoder that maps an image to a $d$-dimensional vector, $g$ is an utterance encoder that maps an utterance to a $d$-dimensional vector, and $\sigma:\R \rightarrow (0,1)$ is the sigmoid function. $d$ is a hyperparameter of our choosing; we choose $d=1024$. $f_L$ is a convolutional neural network (CNN; \shortciteauthor{lecun1995convolutional}, \citeyearNP{lecun1995convolutional}); $g$ is a unidirectional Gated Recurrent Unit (GRU; \shortciteauthor{cho2014learning}, \citeyearNP{cho2014learning}).

When training $\L$, three training examples are extracted from each reference game. The ground-truth referring expression is paired up with each of the three referents: the expression paired with the target forms a positive example, and the other two pairs form negative examples.

Both the literal speaker and the RSA speaker are evaluated on the performance of this trained evaluation listener.

\paragraph{RSA speaker} 
The RSA speaker $\SRSA$ chooses a referring expression by recursively reasoning about a simulated listener's interpretation. The speaker does so by considering how each utterance would be interpreted by an internal listener model $L_{RSA}$, which shares the same formulation as $L_\text{Eval}$ (Eq.~\ref{eq:listener}). After considering each possible utterance, the speaker chooses the most effective and efficient utterance, i.e., the one with the highest \emph{utility}. 
An utterance's utility $U$ is defined as a trade-off between maximizing the likelihood of the listener identifying the correct target $r_t$ and minimizing the cost $C(u)$ of producing utterance $u$: 
\begin{equation} \label{eq:utility}
   U(u|t, r_1, r_2, r_3) = \log L_\text{RSA} (r_t \mid u, r_1, r_2, r_3) - C(u).
\end{equation}
We define $C(u) := \lambda |u|$ to be a linear penalty on the length of an utterance, where $|u|$ denotes the number of words $u$ contains. We choose $\lambda = 0.01$. 

Once $\SRSA$ has computed the utility for each possible utterance, it produces the utterance that maximizes the utility: 
\begin{equation} \label{eq:u_prod}
    \SRSA(t, r_1, r_2, r_3) = {\arg\max}_{u\in \U} U(u|t, r_1, r_2, r_3).
\end{equation}

$L_\text{RSA}$ also learns semantics from data, encapsulated in its own semantic function $\L$. The learned nature of the semantic function introduces the potential for variation. To improve the robustness of $L_\text{RSA}$, we follow \shortciteA{wang2021calibrate} and model $\L$ as the mean of an ensemble of $n$ semantic functions $\{\L^{(1)}, \cdots, \L^{(n)}\}$. All models in the ensemble are identical in architecture to $\L_\text{Eval}$. We choose $n=9$.

\paragraph{Literal speaker with context}
The literal speaker with context $\SLit$ is an image-to-text model with an encoder--decoder architecture. The image encoder $f_S$ produces a vector representation of the context, i.e., the three referents. The decoder $\ell$ takes as input the output of $f_S$ along with the target index to generate an utterance. More specifically, for a given reference game with referents $r_1, r_2, r_3$, and a target index $t \in \{1, 2, 3\}$, $\SLit$ obtains an embedding of the reference game, $\vec{h}$, by first encoding the three referents individually with $f_S$ and concatenating the encodings with the target index $t$:
\begin{equation}
    \vec{h}(t, r_1, r_2, r_3) = [f_S(r_1); f_S(r_2); f_S(r_3); t]
\end{equation}
Then, $\ell$ receives $\vec{h}$ as input and generates an utterance:
\begin{equation}
    \SLit(t, r_1, r_2, r_3) = \ell(\vec{h}(t, r_1, r_2, r_3))
\end{equation}

Here, $f_S$ is an image encoder with the same architecture as $f_L$, and $\ell$ is a GRU trained using the autoregressive language modeling objective with teacher forcing, which uses cross-entropy loss to penalize deviations from the ground-truth.

\paragraph{Training and implementation}
We adapted the neural module implementations of \shortciteA{white2020learning}. Each model was trained for $100$ epochs with a batch size of $32$, and optimized using Adam \shortcite{kingma2014adam} with a learning rate of $0.001$ for the literal speaker and $0.01$ for the semantic functions. For the literal speaker, we used validation accuracy as the model selection criterion. For the semantic functions, we used validation loss as the model selection criterion.\footnote{All code and analyses for this paper can be found at \url{github.com/feifang24/overmod-from-pragmatic-learning}.}

\section{Experiment 1}

We first investigate the degree to which the literal and RSA speakers choose overinformative referring expressions when all colors are equally likely to occur with all objects. This imitates the distribution of non-color-diagnostic objects, for which color overmodification is common 
\shortcite{sedivy2003pragmatic}.

\paragraph{Setup}
We created a dataset of $75,000$ reference games, in which all color--shape combinations appear at a uniform frequency.
$55,000$ ($\approx 73\%$) of these were used to train the neural agents, and the remaining were reserved for evaluation.

The reference games used for training were split into $11$ subsets of equal size: one for training the literal speaker $\SLit$, one for training the semantic function $\L$ in $L_{Eval}$, and nine for training the ensemble of semantic functions in $\SRSA$'s internal listener $L_\text{RSA}$. The games were generated from the same distribution, but no two models were trained on identical data.

\begin{figure}

    \begin{minipage}{0.015 \textwidth}
        \includegraphics[scale=0.3]{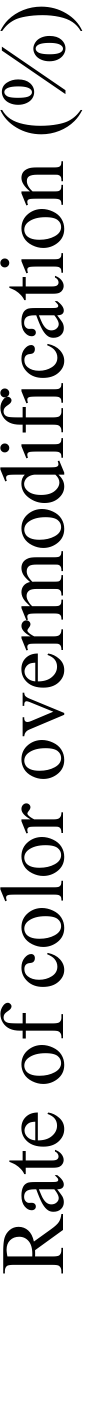}
    \end{minipage}
    \begin{minipage}{0.11 \textwidth}
        \includegraphics[width=0.89\textwidth]{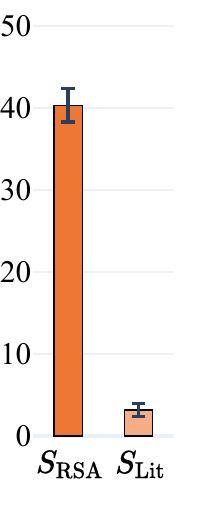}
    \end{minipage}
    \begin{minipage}{0.015 \textwidth}
        \includegraphics[scale=0.3]{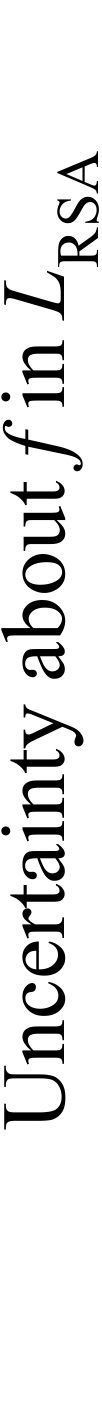}
    \end{minipage}
    \begin{minipage}{0.3 \textwidth}
        \includegraphics[width=\textwidth]{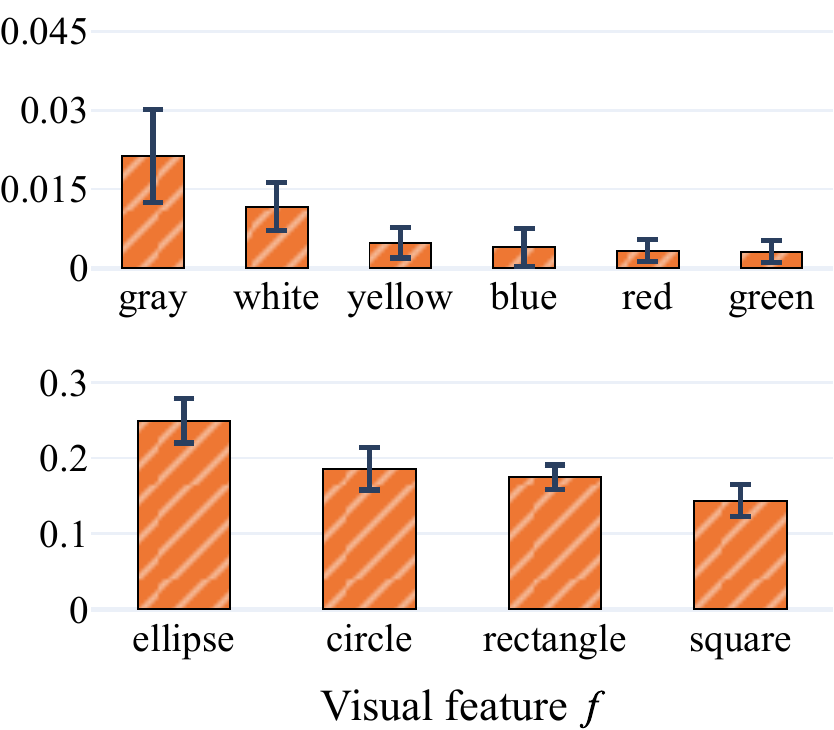}
    \end{minipage}
    \caption{Results of Exp.~1 showing speakers' rates of color overmodification (left), and the uncertainty associated with all visual features in the internal listener model $L_\text{RSA}$ (right). Error bars correspond to 95\% confidence intervals across 5 random initializations during training.}
    \label{fig:baseline-results}
\end{figure}

\paragraph{Results}
We start by noting that both speaker models have successfully learned to communicate the target to the evaluation listener with high accuracy ($\SLit$: 93\%, $S_{RSA}$: 94\%). With this, we can turn to inspect the overmodification strategies the speaker models employ to achieve this accuracy.

To investigate speakers' color overmodification behavior, we focus on the rate of color in addition to shape mention (e.g., ``blue square'') when color is neither sufficient nor necessary to identify the target (i.e., shape-needed condition in Figure~\ref{fig:games-by-condition}). While strictly unnecessary, both speakers use color to overmodify, but the RSA speaker does so at a much higher rate (see Figure~\ref{fig:baseline-results}, left). The results are borne out as a significant main effect of the speaker on the rate of overmodification in a linear regression model ($\beta = .37$, $\text{SE} = .01$, $p<.001$). This suggests that overmodification with color is specifically encouraged in speakers that explicitly reason about listeners.\footnote{Future work should explore how the rate of overmodification in these models is further affected by increased scene variation (e.g., \shortciteauthor{koolen2011factors}, \citeyearNP{koolen2011factors}).} 

While the literal speaker remains a black box, we can inspect the RSA speaker's internal listeners to investigate what gives rise to the rate of color overmodification. More specifically, we can look at which visual features of the target are most or least likely to prompt the speaker to overmodify, and relate these patterns to the listener's understanding of these features. We define \textit{uncertainty} in the listener's interpretation as the extent to which the outputs of the listener's probabilistic semantic function $\L$ deviate from the outputs from a truth-conditional semantics, which are 1 when the utterance applies to a referent and 0 otherwise.\footnote{Treating deviation from end points as uncertainty is consistent with the probabilistic interpretation of real-valued semantics in \shortcite{degen2020redundancy}, as well as with the classifier training objective for the literal listeners.}
Figure \ref{fig:baseline-results} (right) shows substantial variation in feature uncertainty, despite balanced training data---a finding we return to in Exp.~3.

In alignment with prior work, we find that overmodification rates are highest where there is high uncertainty for shape and low uncertainty for color \shortcite{degen2020redundancy}. $\SRSA$ overmodifies with color most frequently when the target is an ellipse, which is associated with the highest uncertainty among all shapes (Figure \ref{fig:baseline-results} bottom right). $\SRSA$ overmodifies with color least frequently when the target is gray, which is associated with the highest uncertainty among all colors (Figure \ref{fig:baseline-results} top right). This observation is supported by a significant interaction of color and shape uncertainty on the rate of overmodification using a linear regression model ($\beta = -24.47$, $\text{SE} = 10.26$, $p<.05$)

Finally, we observe that when both color and shape are necessary for unique identification of the target (\emph{both needed} condition, Figure \ref{fig:games-by-condition}), $\SRSA$ and $\SLit$ achieve comparable listener accuracy ($90.43\pm 0.20\% $ and $90.31\pm 0.56\%$, respectively). However, when color is strictly not necessary (\emph{shape needed}) but $\SRSA$ overmodifies to a high degree, $\SRSA$ achieves a higher listener accuracy than $\SLit$ ($87.56\pm 0.34\%$ and $85.32\pm 0.97\%$, respectively), borne out by a significant interaction between speaker and context condition ($\beta = .02$, $\text{SE} = .01$, $p<.01$). This suggests that overmodification might be an effective strategy that the RSA speaker leverages to improve communicative accuracy.

    \begin{figure}
    \centering
        \begin{minipage}{0.04 \textwidth}
            \includegraphics[scale=0.3]{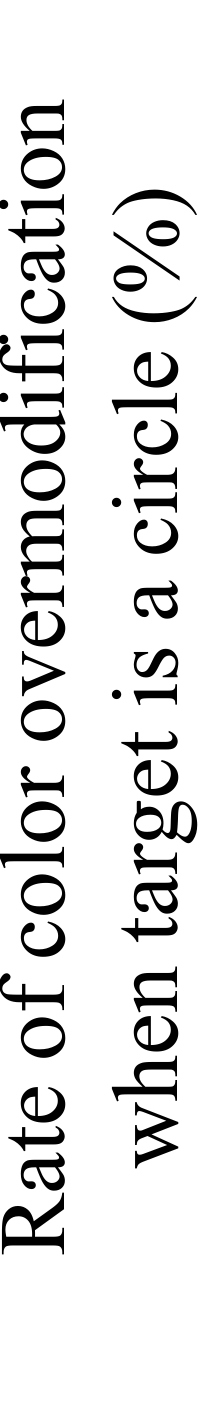}
        \end{minipage}\hfill
        \begin{minipage}{0.44 \textwidth}
            \begin{minipage}{ \textwidth}
            \includegraphics[scale=0.35, right]{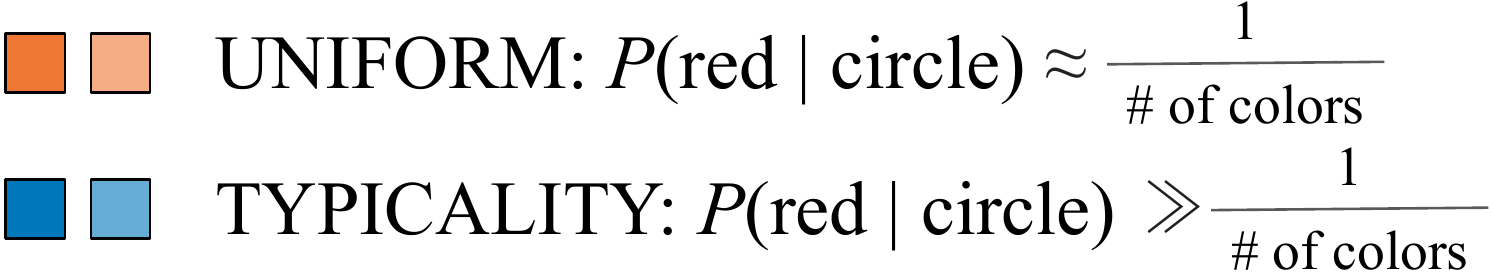}
            \end{minipage}
                \begin{minipage}{0.48 \textwidth} \includegraphics[scale=0.475]{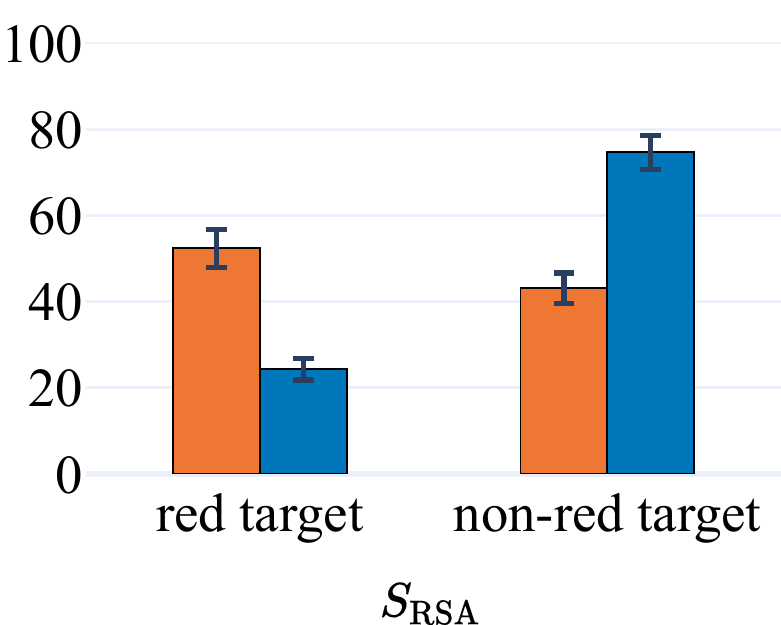}
                \end{minipage}
                \begin{minipage}{0.48 \textwidth} \includegraphics[scale=0.475]{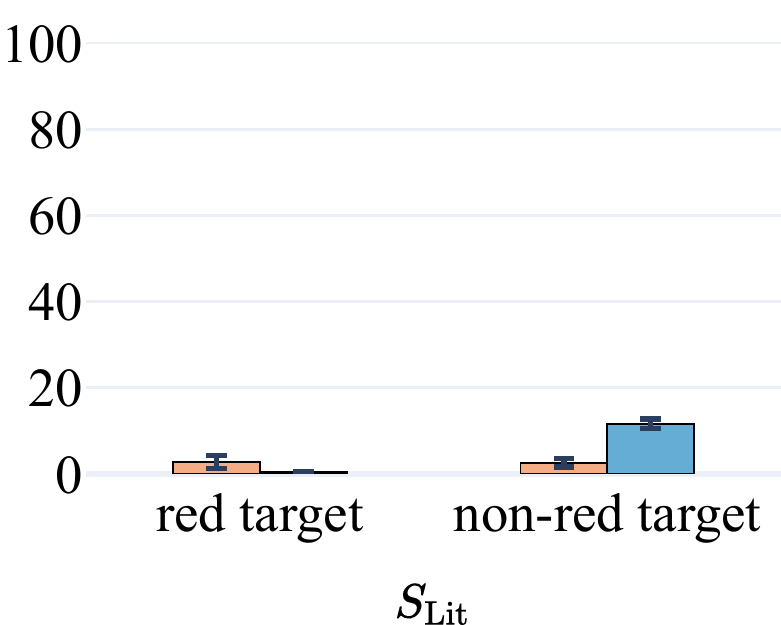}
                \end{minipage}
        \end{minipage}
        \caption{Rates of color overmodification when referring to red and non-red circles, after learning from uniformly colored circles (Exp.~1) and circles that are typically red (Exp.~2).}
        \label{fig:typicality-overmod}
    \end{figure}

\section{Experiment 2}
Exp.~1 establishes that both the RSA speaker and the literal speaker generate referring expressions that overmodify with color, although at different rates. The speaker models were trained on underlying data where each color was uniformly paired with each shape, which is a commonly used feature distribution for models trained on synthetic data (e.g., ShapeWorld \shortcite{kuhnle2017shapeworld} or CLEVR \shortcite{johnson2017clevr}). Conceptually, the uniform distribution of features makes color non-diagnostic for all object types \shortcite{tanaka1999diagnosticity}. In our second experiment, we investigate how the speaker models' overmodification changes when certain objects are more likely to occur in some colors than in others. This setup imitates the distribution of color-diagnosticity in the real world; humans refer to such objects overinformatively when they are atypically-colored but less so when typically colored \shortcite{westerbeek2015typicality, sedivy2003pragmatic, degen2020redundancy, kreiss2020production}.

\paragraph{Setup}
    
We repeat the experimental protocol from Exp.~1 with a new dataset, \textsc{typicality}, in which the frequency of certain color--shape combinations deviates from uniform. Specifically, 
90\% of all target circles are red, making circles in any other color atypical. We then examine how the overmodification behavior of a speaker trained on \textsc{typicality} differs from that of a speaker trained on the dataset from Exp.~1, which will constitute the \textsc{uniform} condition.

\paragraph{Results}
    
\begin{figure}
    
    \begin{minipage}{0.5\textwidth}
    \centering
        \includegraphics[scale=0.22]{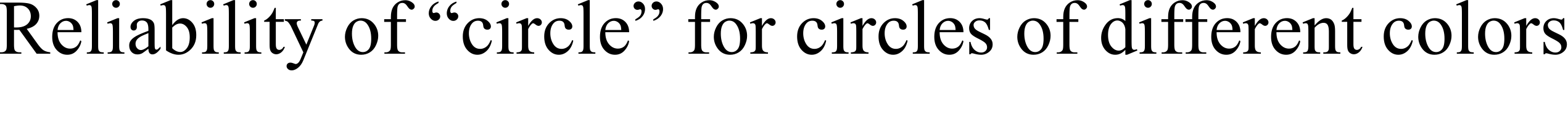}
    \end{minipage}
    
    \begin{minipage}{0.5\textwidth}
    \centering
        \includegraphics[scale=0.65]{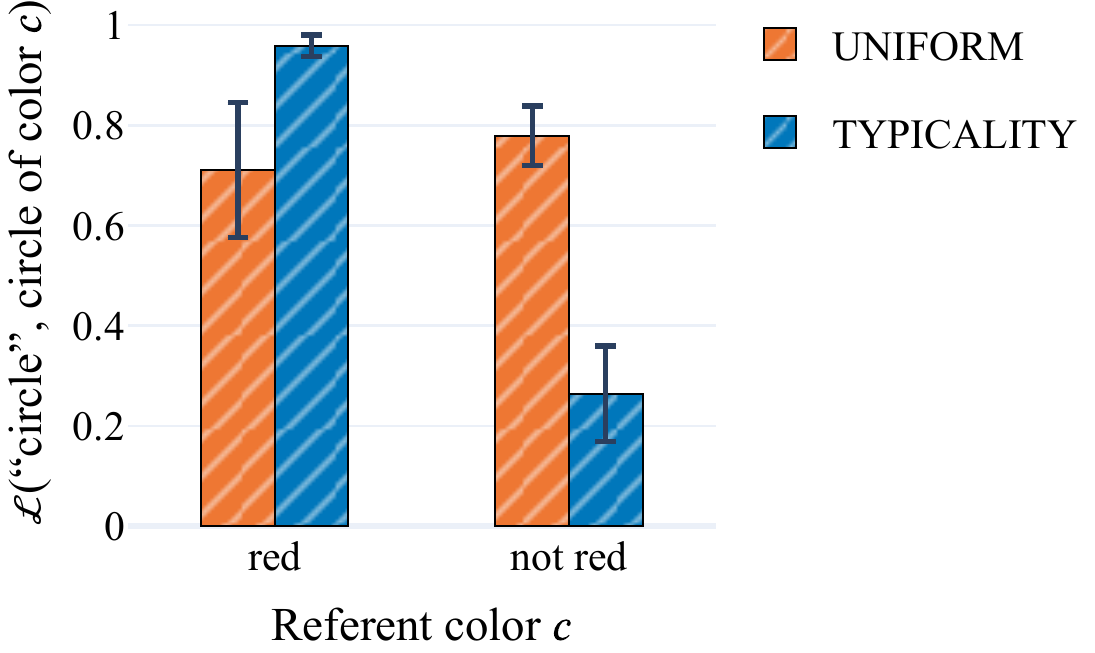}
    \end{minipage}
    \caption{The internal listener model $L_\text{RSA}$'s judgment about how well the utterance ``circle'' applies to circles of different colors, after learning from \textsc{uniform} and \textsc{typicality}. }
    \label{fig:typicality-sem}
\end{figure}
    
Figure~\ref{fig:typicality-overmod} shows how the data distribution impacts each speaker's frequency of color overmodification when referring to a circle. Qualitatively, both speakers' rates of overmodification decrease as the color becomes more typical (red in this case) and vice versa, resembling the empirical findings from prior literature (e.g., \shortciteauthor{westerbeek2015typicality}, \citeyearNP{westerbeek2015typicality}). However, as in Exp.~1, $\SRSA$ overmodifies significantly more frequently than $\SLit$, regardless of object typicality. These observations are borne out in a linear regression analysis as main effects of speaker ($S_{RSA}$ vs.~$\SLit$; $\beta = .63$, $\text{SE} = .02$, $p<.001$) and color of the circle (red vs.~non-red; $\beta = -.11$, $\text{SE} = .02$, $p<.001$), as well as a significant interaction ($\beta = -.39$, $\text{SE} = .02$, $p<.001$).

Considering the range of color overmodification rates in the empirical typicality literature (e.g., typical: $\approx 20\%$; atypical: $\approx 75\%$ in \shortciteA{westerbeek2015typicality}, and typical: $\approx 25\%$; atypical: $\approx 55\%$ in \shortciteA{degen2020redundancy}), the color production rates of $\SRSA$ are also quantitatively expected, whereas $\SLit$ underpredicts those rates. 

We now again turn to investigate what gives rise to the overmodification behavior in $\SRSA$ by examining its internal listeners' semantic functions, which represent the listeners' judgment about how well an utterance applies to a referent. When a listener is trained on \textsc{typicality}, the listener believes that ``circle'' applies well to a circle when it is of a typical color (i.e., red) and poorly otherwise (Figure~\ref{fig:typicality-sem}). In contrast, when the listener is trained on the \textsc{uniform} data, the listener perceives circles of all colors similarly. 
In other words, the listener's semantic function significantly changes with the color--shape distribution it is exposed to during learning. This is borne out in a linear model as a significant interaction between dataset and red vs. non-red circles ($\beta = -.76$, $SE = .03$, $p<.001$). The results suggest that the overmodification asymmetry observed as a typicality effect can be captured simply by exposing a learning pragmatic agent with a non-uniform color distribution. 

The results of the listener investigation are in line with the assumed semantics in prior work where the unmodified utterance (e.g., ``banana'') is mapped with higher certainty to typically colored objects (yellow bananas) than to atypically colored ones (blue bananas) \shortcite{degen2020redundancy}. 

In summary, Exp.~2 shows that properties of the underlying data distribution play an important role in a speaker's patterns of overmodification. While the overall rate in $\SRSA$ is more compatible with previously observed human data, both speakers exhibit the same qualitative pattern of overmodification.

\begin{figure}
    \centering
    \includegraphics[width=0.45\textwidth]{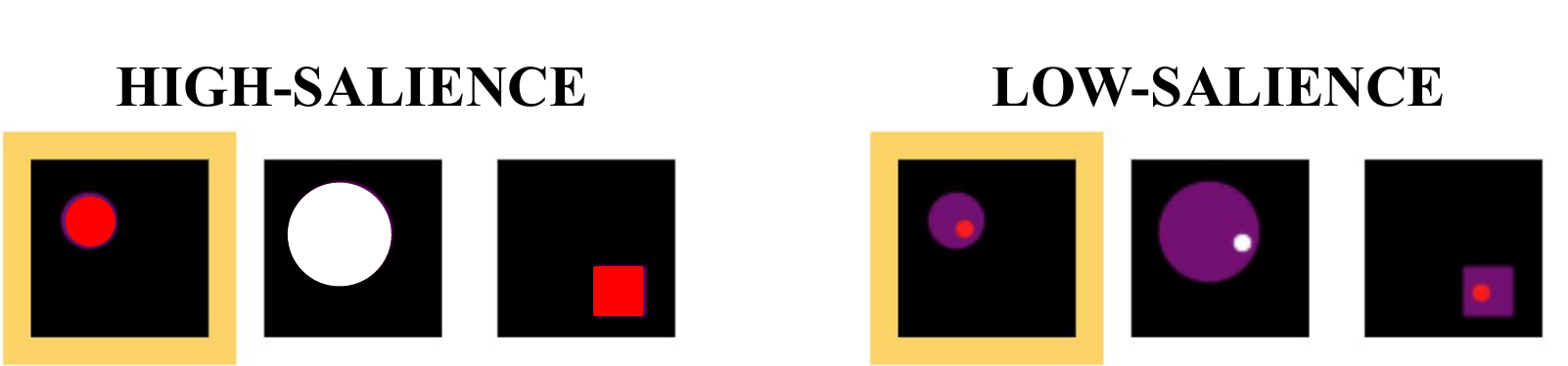}
    \caption{Example reference games from the \textsc{high-salience} (Exp.~1) and \textsc{low-salience} conditions, respectively, both with the ground-truth utterance ``red circle.''}
    \label{fig:low-salience-demo}
\end{figure}

\section{Experiment 3}
The previous experiments explored whether color overmodification patterns from prior literature naturally emerge from neural speakers learning in certain environments where feature frequencies vary. The structured manipulation of the learning data led to changes in the learned uncertainties over features, which gave rise to the expected overmodification patterns. 
However, variation in uncertainty cannot simply be reduced to differences in frequency (e.g., see the uncertainty variation between \emph{gray} and \emph{red} in Figure~\ref{fig:baseline-results}). We now explore how overmodification changes when we leave the overall data frequency constant but make color more difficult to identify, rendering it less salient\footnote{Salience is generally underspecified. Here, we take a simplistic approach and operationalize salience as ease of feature extraction.}. We thus generated a novel hypothetical world where an object's color is determined by only a single colored pixel (right context in Figure~\ref{fig:low-salience-demo}). 
Following prior work, we expect that the reduced salience should reduce overinformative color use \shortcite{davies2013speakers,sedivy2003pragmatic,rubio2016redundant,van2019conceptualization}.

\paragraph{Setup}

We use the dataset from Exp.~1 as the \textsc{high-salience} condition and construct a novel variant of it, the \textsc{low-salience} condition (see Figure~\ref{fig:low-salience-demo}). While a shape's color in the high-salience condition can be determined by randomly sampling \emph{any} pixel within the shape, color in the low-salience condition needs to be extracted from one specific single pixel that is randomly placed within the shape. 

    \begin{figure}
        \centering
        \begin{minipage}{0.33 \textwidth}
        \includegraphics[scale=0.72]{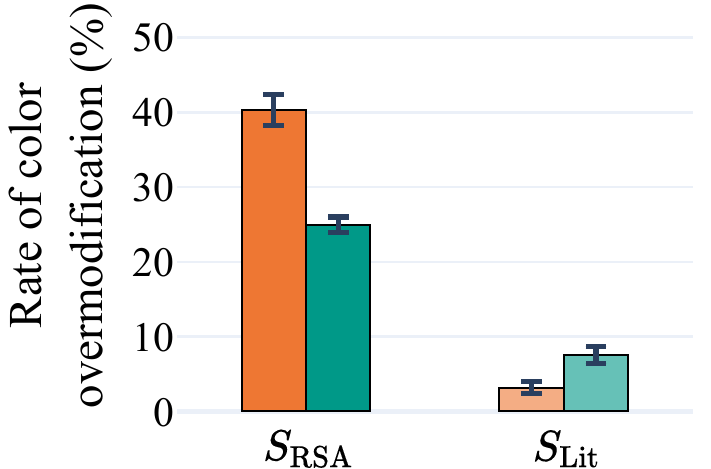}
        \end{minipage}
        \begin{minipage}{0.13 \textwidth}
        \includegraphics[scale=0.3]{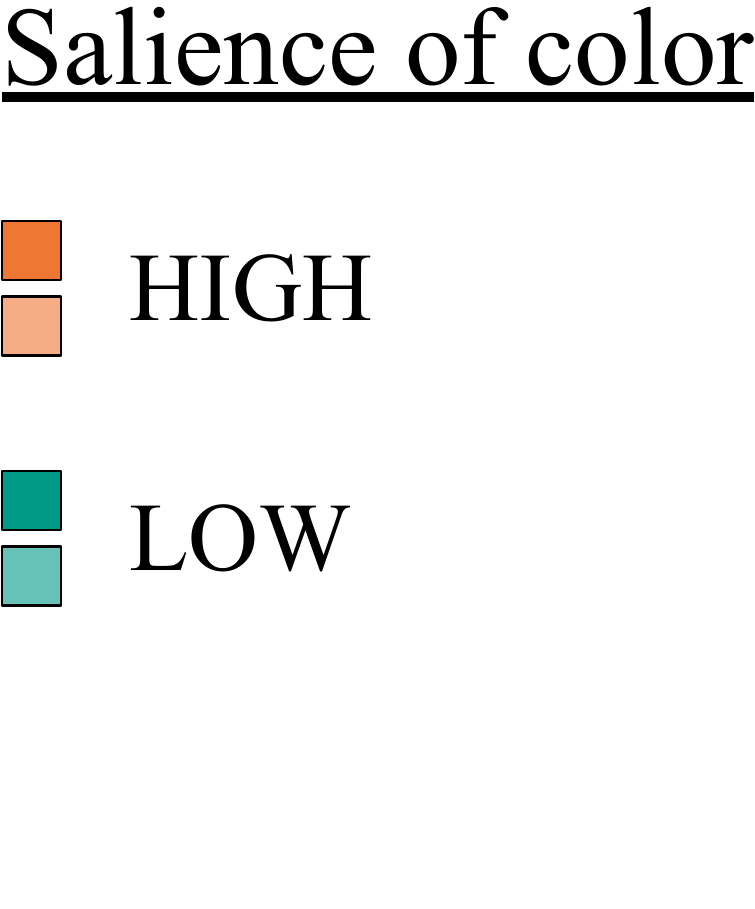}
        \end{minipage}
        \caption{Frequency of color overmodification in the two speakers, trained on \textsc{high-salience} (Exp.~1) and \textsc{low-salience} (Exp.~3), respectively.}
        \label{fig:salience-overmod}
    \end{figure}

\paragraph{Results}
As seen across experiments, $\SRSA$ overmodifies with color more frequently than $\SLit$ regardless of the salience of color in the data distribution they learn from (Figure~\ref{fig:salience-overmod}). However, in this experiment, $\SRSA$ and $\SLit$ even come apart qualitatively. As color becomes less salient, $\SRSA$’s frequency of color overmodification drops, while $\SLit$’s increases. These observations are borne out in a linear regression analysis with main effects of speaker ($\beta = .37$, $\text{SE} = .01$, $p<.001$), salience ($\beta = .04$, $\text{SE} = .01$, $p<.001$), as well as their interaction ($\beta = -.20$, $\text{SE} = .01$, $p<.001$).

The behavior of $\SRSA$ is consistent with prior work suggesting that speakers might especially overmodify with color because of its salient nature \shortcite{van2019conceptualization,rubio2016redundant,davies2013speakers,sedivy2003pragmatic}. While $\SRSA$'s overmodification patterns are aligned with prior work, $\SLit$'s are at odds with it.

Upon inspecting the $\SRSA$'s internal listener interpretations, we see that color modifiers are associated with about $10$ times more uncertainty when the listener learns from an environment with low color salience ($0.094 \pm 0.063$) than from one with high color salience ($0.008 \pm 0.002$). This difference is borne out as a main effect in a linear model ($\beta = .09$, $\text{SE} = .01$, $p<.001$). 
The results are consistent with the findings from Exp.~1, which show that the RSA speaker overmodifies infrequently when the redundant information is associated with high uncertainty \shortcite{degen2020redundancy}.

\section{Conclusion}
When referring to objects in the world, speakers often choose to mention an object's color even when it is not necessary for uniquely picking it out among competitors (e.g., \shortciteauthor{pechmann1989overspecification}, \citeyearNP{pechmann1989overspecification}). We have shown that a range of these attested overmodification patterns emerge naturally in neural speaker agents that learn semantics from data and reason pragmatically about internal listeners. Our findings align with color overmodification patterns reported for human speakers, and help to deepen existing modeling results concerning these patterns. More generally, we hope to have shown that neural networks are powerful tools for exploring how learning in different environments can shape pragmatic language use.

\newpage

\bibliographystyle{apacite}

\setlength{\bibleftmargin}{.125in}
\setlength{\bibindent}{-\bibleftmargin}

\bibliography{cogsci}

\end{document}